\documentclass[letterpaper, 10 pt, conference]{ieeeconf}  

\IEEEoverridecommandlockouts                              

\usepackage{amsmath} 
\usepackage{amssymb}  
\usepackage{ascmac}
\usepackage{xcolor}
\usepackage{graphicx}
\usepackage{booktabs}
\usepackage{cite}
\usepackage{algorithm}
\usepackage{algorithmic}
\usepackage{multirow}
\usepackage{subcaption}
\usepackage{url}
\usepackage{hyperref}

\usepackage{eso-pic} 

\title{CoLF: Learning Consistent Leader–Follower Policies for Vision-Language-Guided  
Multi-Robot Cooperative Transport \\
}

\author{Joachim Yann Despature$^{1,2}$, Kazuki Shibata$^{1}$ and Takamitsu Matsubara$^{1}$
\thanks{$^{1}$All the authors are with the Division of Information Science, Graduate School of Science and Technology, Nara Institute of Science and Technology (NAIST), Nara, Japan.}
\thanks{$^{2}$Joachim Yann Despature is with École polytechnique fédérale de Lausanne (EPFL), Lausanne, Switzerland.}
}

\begin{document}

\AddToShipoutPicture*{%
    \AtPageUpperLeft{%
        \raisebox{-1.2cm}{ 
            \hspace{1cm} 
            \parbox{\textwidth}{\centering \small 
                This work has been submitted to the IEEE for possible publication. Copyright may be transferred without notice, after which this version may no longer be accessible.
            }
        }
    }
}

\maketitle
\thispagestyle{empty}
\pagestyle{empty}

\begin{abstract}
In this study, we address vision--language-guided multi-robot cooperative transport, where each robot grounds natural-language instructions from onboard camera observations. A key challenge in this decentralized setting is perceptual misalignment across robots, where viewpoint differences and language ambiguity can yield inconsistent interpretations and degrade cooperative transport. To mitigate this problem, we adopt a dependent leader--follower design, where one robot serves as the leader and the other as the follower. Although such a leader--follower structure appears straightforward, learning with independent and symmetric agents often yields symmetric or unstable behaviors without explicit inductive biases. To address this challenge, we propose Consistent Leader--Follower (CoLF), a multi-agent reinforcement learning (MARL) framework for stable leader--follower role differentiation. CoLF consists of two key components: (1) an asymmetric policy design that induces leader--follower role differentiation, and (2) a mutual-information-based training objective that maximizes a variational lower bound, encouraging the follower to predict the leader’s action from its local observation. The leader and follower policies are jointly optimized under the centralized training and decentralized execution (CTDE) framework to balance task execution and consistent cooperative behaviors. We validate CoLF in both simulation and real-robot experiments using two quadruped robots. The demonstration video is available at https://sites.google.com/view/colf/.
\end{abstract}

\section{Introdution}
\label{section:introduction}
This study addresses vision--language-guided multi-robot cooperative transport, in which robots cooperatively transport a target object toward a goal landmark specified by a natural-language instruction. Recent advances in Vision--Language Models (VLMs) have enabled visual grounding of natural-language instructions, making them a promising tool for multi-robot control. Most existing approaches adopt centralized architectures that ensure consistent interpretation across robots using global observations~\cite{CollaBot,VIKI-R,COMBO,HierarchicalLM,TRACE,MCoCoNav,ZeroCAP,CLIPSwarm}. However, such approaches rely on global-view sensing and extensive inter-robot communication, which limits their applicability in real-world environments. This limitation motivates decentralized approaches in which each robot performs VLM inference independently from onboard observations, without relying on global-view sensors or extensive communication.

A key challenge in this decentralized setting is perceptual misalignment across robots, where different robots associate the same instruction or environment with different perceived objects or goal regions. Such misalignment undermines coordinated behavior, as inconsistent interpretations directly induce conflicting actions during cooperative transport. This issue is particularly critical in multi-agent systems composed of independent and symmetric agents, each equipped with its own perception and decision-making modules.

\begin{figure}[t]
  \centering
  \includegraphics[width=0.9\linewidth]{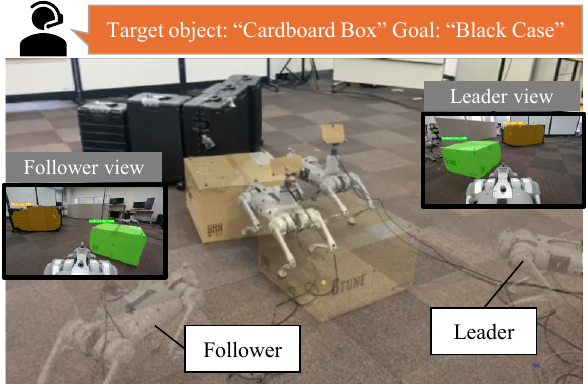}
  \caption{Overview of vision--language-guided multi-robot cooperative transport. Two robots cooperatively transport the target object toward a language-specified goal landmark using onboard RGB-D observations.}
  \label{fig:pull}
\end{figure}

To mitigate this issue, we depart from the conventional independent and symmetric multi-agent paradigm by adopting a dependent leader--follower design~\cite{Wang2016}, in which one robot acts as the leader and the other as the follower. By centralizing task-relevant interpretation of instruction-referenced elements in the leader, this design mitigates inter-robot perceptual misalignment and reduces implementation costs, as high-accuracy interpretation is required for only a single agent.

Although such a dependent leader--follower structure may appear conceptually straightforward, such role differentiation rarely emerges from naive trial-and-error learning. Without explicit inductive biases, independent policy learning tends to converge to symmetric or unstable behaviors rather than a consistent leader--follower structure.

In this paper, we design and introduce explicit inductive biases to induce a consistent leader--follower structure. Building on this design principle, we propose Consistent Leader--Follower (CoLF), a multi-agent reinforcement learning (MARL) framework for stable leader--follower role differentiation in a two-robot system. CoLF consists of two key components: (1) an asymmetric policy input design that provides the leader with task-level information while restricting the follower to local observations, and (2) a mutual-information-based training objective that maximizes a variational lower bound, which encourages the follower to predict the leader’s action from its local observation. The leader and follower policies are jointly optimized under the centralized training and decentralized execution (CTDE) framework o balance task execution and consistent cooperative behaviors.

Although our approach is inspired by model-based control methods~\cite{Wang2016} in that it employs a leader–follower structure, these model-based methods assume that the robots’ contact points are rigidly attached to the object. In contrast, our MARL-based approach does not rely on an explicit dynamics model, enabling applicability in scenarios where contact points are not rigidly attached. To the best of our knowledge, no prior work combining leader–follower design with MARL has been reported so far.

To evaluate the effectiveness of the proposed method, we conducted cooperative transport experiments with two quadruped robots in simulation and on real hardware. In simulation, we considered one-goal-landmark and two-goal-landmark settings to assess robustness to perceptual misalignment caused by viewpoint differences and ambiguous language instructions. In real-robot experiments, we evaluated the proposed method across five scenarios with different goal landmarks, including cases that induce perceptual misalignment, and confirmed sim-to-real transfer and generalization to various goal landmarks.

The contributions of this paper are summarized as follows:
\begin{itemize}
    \item We propose CoLF, a multi-agent reinforcement learning framework for vision–language-guided multi-robot cooperative transport that operates without global-view sensors or extensive inter-robot communication.
    \item We show through simulation that CoLF consistently outperforms baseline methods under perceptual misalignment caused by viewpoint differences and language ambiguity.
    \item  We validate the proposed approach on real quadruped robots, demonstrating superior performance over the strongest baseline across five cooperative transport scenarios, including challenging cases with perceptual misalignment.
\end{itemize}

\section{Related work}
\subsection{Model-based Multi-robot Cooperative Transport}
Model-based multi-robot cooperative transport relies on a dynamics model to represent robot–object interactions and derives control laws to regulate the object state toward a desired state. It is typically formulated as constrained optimal control~\cite{Hu2022,Koung2021,Kennel2024} or model predictive control (MPC)~\cite{Li2023,Sihao2025}, incorporating friction and tension limits as well as collision and tip-over avoidance. To meet real-time requirements under communication and computational constraints, consensus-based distributed optimization and distributed predictive control, including distributed nonlinear MPC, have also been proposed~\cite{Fawcett2023,Carli2025}. Moreover, decentralized adaptive controllers provide Lyapunov-based guarantees of asymptotic convergence even with unknown object parameters~\cite{Culbertson2021,Yan2021}.

While model-based cooperative transport methods offer theoretical guarantees on stability and convergence, they typically assume that the robots’ contact points are rigidly attached to the object to formulate a dynamics model. In contrast, our method is model-free and does not assume an explicit dynamics model, enabling applicability in scenarios where contact points are not rigidly attached.

\subsection{MARL-based Multi-robot Cooperative Transport}
Several recent studies on multi-robot cooperative transport have adopted MARL without relying on a dynamics model. These approaches have been applied to a wide range of robotic platforms, quadrupedal robots for cooperative pushing~\cite{Feng2025}, bipedal robots for cooperative transport~\cite{Pandit2024,Mehta2025}, and aerial robots for cable-suspended load transport~\cite{Zeng2025}.

A common design choice in MARL-based cooperative transport is the CTDE paradigm: global information is exploited during training to mitigate non-stationarity, whereas each robot selects actions during execution using a decentralized policy based on local observations. Furthermore, several recent studies focus on scalability to team-size changes, including consensus-based estimation in distributed policies~\cite{Shibata2023RAS}, curriculum learning with knowledge distillation for improved generalization~\cite{Chen2025}.

While these MARL-based cooperative transport methods enable multi-robot cooperation without relying on an explicit dynamics model, they typically assume that the object and goal landmark poses are available during execution. In contrast, our approach leverages VLM-based inference and does not require precise pose coordinates, enabling vision-language-guided multi-robot cooperative transport.

\subsection{VLM-based Multi-robot Control}
VLM-based multi-robot control has emerged as a promising approach for enabling robots to interpret task specifications and execute cooperative tasks. By grounding natural-language instructions in visual observations, VLMs have been applied to a wide range of multi-robot cooperative tasks, including cooperative manipulation~\cite{CollaBot}, home-service tasks~\cite{VIKI-R,COMBO,HierarchicalLM}, cooperative transport and navigation~\cite{TRACE}, cooperative exploration~\cite{MCoCoNav}, and formation control~\cite{ZeroCAP,CLIPSwarm}.

Many existing studies on VLM-based multi-robot control adopt centralized architectures that leverage a global observation for VLM inference. One line of work uses a single top-down image to infer target configurations or interpret instructions~\cite{ZeroCAP,CLIPSwarm}. Another line of work fuses robot views into a shared map or global scene representation for centralized subgoal assignment and planning~\cite{COMBO,VIKI-R,HierarchicalLM}. Recent work has also explored a distributed approach, in which each robot performs VLM inference locally from its own observations while relying on a shared global semantic map~\cite{MCoCoNav}.

While these studies facilitate high-level task interpretation and multi-robot coordination, they often rely on global-view sensors or extensive inter-robot communication, which can impose practical constraints in real-world scenarios. In contrast, our approach assumes neither global-view sensors nor explicit inter-robot communication by leveraging a dependent leader–follower policy structure with decentralized execution. This enables decentralized cooperative transport using each robot's local observations and the given language instruction.

\section{Preliminary}
\subsection{Vision-language-guided Multi-robot Cooperative Transport}
In this study, we consider cooperative pushing with two quadrupedal robots, where each robot executes a decentralized policy based on onboard visual observations and a natural-language instruction that specifies the target object and a goal landmark.
The control objective is to cooperatively transport the specified object to the goal landmark specified by a language instruction.

We formulate the problem under the following assumptions:
\begin{itemize}
  \item During execution, each robot does not have access to the coordinates of the target object and the goal landmark.
  \item For collision avoidance, each robot can obtain the relative position and relative yaw angle of the other robot.
  \item The leader correctly recognizes the target object and the goal landmark during execution.
\end{itemize}

The other robot’s relative position and yaw angle can also be inferred from visual observations; however, reliable estimation typically requires viewpoint control to maintain visibility of the other robot while pushing the object. To isolate this problem, we treat this information as available.

\subsection{Decentralized Partially Observable Markov Decision Process (Dec-POMDP)}
In this study, we formulate the multi-robot cooperative transport task as a Dec\mbox{-}POMDP \cite{Dec-POMDP}.
We define the tuple $\langle \mathcal{S}, \mathcal{B}, \{\mathcal{A}^i\}_{i\in\mathcal{B}}, P, \{r^i\}_{i\in\mathcal{B}}, \{\mathcal{O}^i\}_{i\in\mathcal{B}}, \gamma \rangle$,
where $s\in\mathcal{S}$ denotes the global state and $\mathcal{B}:=\{1,\dots,n\}$ denotes the set of robots.
At each timestep $t$, robot $i$ receives a local observation $o_t^i$ and selects an action $a_t^i$, forming the joint action $a_t = (a_t^1,\dots,a_t^n)$.
The environment transitions from $s_t$ to $s_{t+1}$ according to $P(s_{t+1}\mid s_t,a_t)$ and returns an individual reward $r_t^i$ to each robot.

The learning objective is to maximize the expected discounted cumulative reward of all robots.
Specifically, we define
$J=\mathbb{E}\!\left[\sum_{i=1}^{n} R_t^i\right]$,
where $R_t^i=\sum_{k=0}^{\infty}\gamma^{k} r_{t+k}^i$ denotes the discounted reward of robot $i$.

\subsection{Multi-Agent Proximal Policy Optimization (MAPPO)}
MAPPO~\cite{mappo} is an actor--critic algorithm under the CTDE paradigm.
During training, a centralized critic estimates the state-value function from the global state of all robots, whereas during execution each robot selects its actions using only local observations.
We denote the state-value function by $V_{\phi}(s_t)$ and the policy of robot $i$ by $\pi_{\theta_i}(a^i_t \mid o^i_t)$, where $\theta_i$ and $\phi$ denote the policy parameters of the robot $i$ and the centralized critic parameters, respectively.

Following PPO, MAPPO updates the policy by minimizing the clipped surrogate loss:
\begin{align}
\mathcal{L}_{\mathrm{actor}}(\theta_i)
= \mathbb{E}_t\!\left[
\min\!\Big(
\rho_t^i A_t,\,
\mathrm{clip}(\rho_t^i,\,1-\epsilon,\,1+\epsilon)\,A_t
\Big)
\right],
\end{align}
where $A_t$ is the advantage estimate computed using generalized advantage estimation (GAE)~\cite{ppo}.
The probability ratio between the updated and previous policies is defined as
$\rho_t^i := \pi_{\theta_i}(a_t^i \mid o_t^i) / \pi_{\theta_i^{\mathrm{old}}}(a_t^i \mid o_t^i)$.

Similar to the actor loss, the critic is trained with a clipped value loss:
\begin{align}
\mathcal{L}_{\mathrm{critic}}&(\phi)
= \mathbb{E}_t \Big[
\max \Big(
    (V_{\phi}(s_t) - \hat{R}_t)^2, \nonumber \\
    &\big(
        \mathrm{clip}\!\left(
            V_{\phi}(s_t),\,
            V_{\phi}^{\mathrm{old}}(s_t)-\epsilon,\,
            V_{\phi}^{\mathrm{old}}(s_t)+\epsilon
        \right)
        - \hat{R}_t
    \big)^2
\Big)
\Big],
\end{align}
where $\hat{R}_t := \frac{1}{n}\sum_{i=1}^{n} R_t^i$, $\epsilon>0$ is the clipping parameter, and $V_{\phi}^{\mathrm{old}}$ is the value estimate from the previous policy update.

\section{Method}
\subsection{Overview of the Proposed Framework}
\begin{figure*}[t]
    \centering
    \begin{subfigure}{0.36\linewidth}
        \centering
        \includegraphics[width=\linewidth]{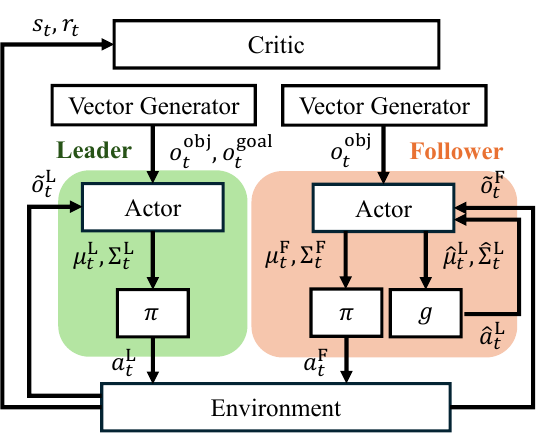}
        \caption{VLM-free training}
        \label{fig:training}
    \end{subfigure}\hfill
    \begin{subfigure}{0.36\linewidth}
        \centering
        \includegraphics[width=\linewidth]{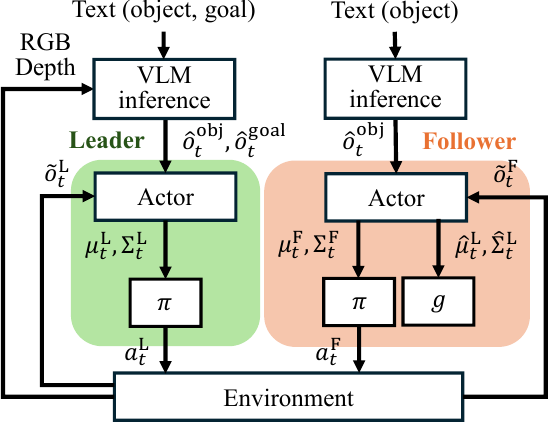}
        \caption{VLM-based execution}
        \label{fig:execution}
    \end{subfigure}\hfill
    \begin{subfigure}{0.20\linewidth}
        \centering
        \includegraphics[width=\linewidth]{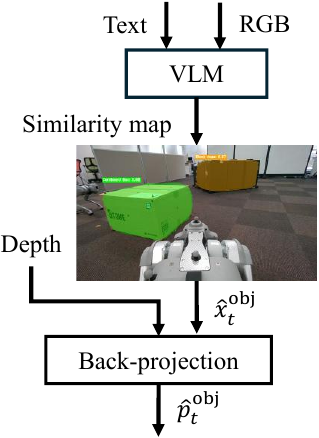}
        \caption{VLM inference}
        \label{fig:VLMinference}
    \end{subfigure}
    \caption{Overview of the CoLF framework for vision-language-guided multi-robot cooperative transport.}
    \label{fig:framework}
\end{figure*}

In this section, we present CoLF, a MARL framework for vision--language-guided multi-robot cooperative transport. An overview of the proposed framework is shown in Fig.~\ref{fig:framework}. CoLF comprises \textit{Leader} and \textit{Follower} agents and explicitly introduces inductive biases toward role acquisition to achieve a consistent leader--follower structure.

CoLF consists of two key components: (1) an asymmetric policy design that induces leader--follower role differentiation, and (2) a mutual-information-based training objective, which encourages the follower to predict the leader’s action from its local observation. To maximize a variational lower bound of this mutual information, we introduce an auxiliary distribution that enables the follower to predict the leader’s action from its local observation. The leader and follower policies are jointly optimized under the CTDE framework to balance task execution and consistent cooperative behaviors.

To reduce computational cost while enabling vision--language-guided cooperative transport at execution time, we adopt VLM-free training and VLM-based execution. This design is inspired by recent LLM-based studies~\cite{Omron2024,Yano2025} that avoid LLM inference during training while leveraging LLMs at execution time. Specifically, the policies are trained using vectorized target object and goal landmark positions, and at execution time they select actions using positions estimated by the VLM.

\subsection{Policy Model}
In MARL with individual VLM inference, viewpoint differences across robots and language ambiguity can lead to inconsistent cooperation, thereby degrading transport performance. 
To address this issue, CoLF introduces an asymmetric policy model with a leader–follower structure to promote consistent cooperative behaviors. 

{\bf Leader policy:}
The leader conditions its policy on the target object and the goal landmark observations. 
We define the leader's local observation as $o_t^{\mathrm{L}} = [o_t^{\mathrm{obj}},\, o_t^{\mathrm{goal}},\,\tilde{o}_t^{\mathrm{L}}]$,
where $o_t^{\mathrm{obj}}$ and $o_t^{\mathrm{goal}}$ denote the observations of the target object and the goal landmark, respectively, and $\tilde{o}_t^{\mathrm{L}}$ contains the remaining local observation. 
The leader actor network $f_{\theta}^{\mathrm{L}}(\cdot)$ parameterizes a Gaussian policy as follows:
\begin{align}
\{\mu_t^{\mathrm{L}}, \Sigma_t^{\mathrm{L}}\}
&= f_{\theta}^{\mathrm{L}}(o_t^{\mathrm{obj}},\, o_t^{\mathrm{goal}},\, \tilde{o}_t^{\mathrm{L}}), \\
\pi(a_t^{\mathrm{L}} \mid o_t^{\mathrm{obj}},\, o_t^{\mathrm{goal}},\, \tilde{o}_t^{\mathrm{L}})
&= \mathcal{N}(\mu_t^{\mathrm{L}}, \Sigma_t^{\mathrm{L}}).
\end{align}
where $\mu_t^{\mathrm{L}}$ and $\Sigma_t^{\mathrm{L}}$ are the mean and covariance.

{\bf Follower policy:}
The follower conditions its policy on the target object observation without the goal landmark.
We define the follower's local observation as $o_t^{\mathrm{F}} = [o_t^{\mathrm{obj}}, \tilde{o}_t^{\mathrm{F}}]$,
where $o_t^{\mathrm{obj}}$ denotes the observation of the target object and $\tilde{o}_t^{\mathrm{F}}$ contains the remaining local observation. 
The follower actor network $f_{\theta}^{\mathrm{F}}(\cdot)$ parameterizes the policy $\pi(a_t^{\mathrm{F}} \mid o_t^{\mathrm{F}})$ and an auxiliary distribution $q(a_t^{\mathrm{L}}\mid o_t^{\mathrm{F}})$ that predicts the leader’s action by
\begin{align}
\{\mu_t^{\mathrm{F}}, \Sigma_t^{\mathrm{F}}, \hat{\mu}_t^{\mathrm{L}}, \hat{\Sigma}_t^{\mathrm{L}}\} &= f_{\theta}^{\mathrm{F}}(o_t^{\mathrm{obj}}, \tilde{o}_t^{\mathrm{F}}), \label{eq:follower_net}\\
\pi(a_t^{\mathrm{F}} \mid o_t^{\mathrm{obj}}, \tilde{o}_t^{\mathrm{F}}) &= \mathcal{N}(\mu_t^{\mathrm{F}}, \Sigma_t^{\mathrm{F}}), \label{eq:follower_pi}\\
g(a_t^{\mathrm{L}} \mid o_t^{\mathrm{obj}}, \tilde{o}_t^{\mathrm{F}}) &= \mathcal{N}(\hat{\mu}_t^{\mathrm{L}}, \hat{\Sigma}_t^{\mathrm{L}}), \label{eq:leader_est}
\end{align}
where $\mu_t^{\mathrm{F}}$ and $\Sigma_t^{\mathrm{F}}$ are the mean and covariance of $\pi(a_t^{\mathrm{F}} \mid o_t^{\mathrm{obj}}, \tilde{o}_t^{\mathrm{F}})$ and $\hat{\mu}_t^{\mathrm{L}}$ and $\hat{\Sigma}_t^{\mathrm{L}}$ are those of $g(a_t^{\mathrm{L}}\mid o_t^{\mathrm{obj}}, \tilde{o}_t^{\mathrm{F}})$.

\subsection{Training Objectives}
\subsubsection{Mutual-Information-Based Objective}
The leader--follower policy is designed to provide the leader with task-level information while restricting the follower to local observations. However, since the follower does not have access to goal information, it may fail to consistently move the target object toward the goal landmark. To address this problem, we introduce the mutual information
$I(a_t^{\mathrm{L}}; o_t^{\mathrm{obj}}, \tilde{o}_t^{\mathrm{F}})$,
which encourages the follower to predict the leader’s action from its local observation and thereby promotes coordinated behavior under the induced information asymmetry.

The mutual information is defined as $I(a_t^{\mathrm{L}}; o_t^{\mathrm{obj}}, \tilde{o}_t^{\mathrm{F}})=H(a_t^{\mathrm{L}})-H(a_t^{\mathrm{L}}\mid o_t^{\mathrm{obj}}, \tilde{o}_t^{\mathrm{F}})$,
where $H(\cdot)$ denotes entropy.
Since $H(a_t^{\mathrm{L}})$ does not depend on the follower parameters,
it is treated as a constant during optimization.
Therefore, the problem reduces to minimizing $H(a_t^{\mathrm{L}} \mid o_t^{\mathrm{obj}}, \tilde{o}_t^{\mathrm{F}})$, defined as
$H(a_t^{\mathrm{L}} \mid o_t^{\mathrm{obj}}, \tilde{o}_t^{\mathrm{F}})
= -\mathbb{E}\!\left[\log p(a_t^{\mathrm{L}} \mid o_t^{\mathrm{obj}}, \tilde{o}_t^{\mathrm{F}})\right]$.

\subsubsection{Policy Optimization}
Minimizing $H(a_t^{\mathrm{L}} \mid o_t^{\mathrm{obj}}, \tilde{o}_t^{\mathrm{F}})$ requires evaluating 
$p(a_t^{\mathrm{L}} \mid o_t^{\mathrm{obj}}, \tilde{o}_t^{\mathrm{F}})$, but it is analytically intractable.
Thus, we introduce a variational distribution
$q(a_t^{\mathrm{L}} \mid o_t^{\mathrm{obj}}, \tilde{o}_t^{\mathrm{F}})$ to approximate $p(a_t^{\mathrm{L}} \mid o_t^{\mathrm{obj}}, \tilde{o}_t^{\mathrm{F}})$.

Using the non-negativity of the KL divergence, we obtain an upper bound on the conditional entropy as
$H(a_t^{\mathrm{L}} \mid o_t^{\mathrm{obj}}, \tilde{o}_t^{\mathrm{F}})
\le - \mathbb{E}\!\left[\log q(a_t^{\mathrm{L}} \mid o_t^{\mathrm{obj}}, \tilde{o}_t^{\mathrm{F}})\right]$.
Accordingly, we introduce the consistency-enhancing loss given by
\begin{equation}
\mathcal{L}_{\mathrm{CE}}
= - \mathbb{E}\!\left[\log q(a_t^{\mathrm{L}} \mid o_t^{\mathrm{obj}}, \tilde{o}_t^{\mathrm{F}})\right].
\label{eq:CEloss}
\end{equation}

During training, the leader’s actor network is updated by minimizing $\mathcal{L}_{\mathrm{actor}}$, whereas the follower’s actor network is updated by minimizing $\mathcal{L}_{\mathrm{actor}} + \lambda \mathcal{L}_{\mathrm{CE}}$, where $\lambda > 0$ is a weighting coefficient for the CE loss. The critic parameters $\phi$ are updated by minimizing $\mathcal{L}_{\mathrm{critic}}$.

\subsection{VLM-free Training and VLM-based Execution}
In MARL with VLMs, training policies across diverse objects and goal landmarks incurs substantial training time, and including VLM inference in the training loop further increases computational cost. To address this issue, we introduce a learning framework that trains policies without VLMs and executes decentralized policies conditioned on VLM estmates.

We train policies using the object position, object velocity, and the goal-landmark position. Since VLM-based velocity estimates are noisy at execution, we adopt an asymmetric actor--critic architecture~\cite{asymmetric-actor-critic}: the critic is conditioned on object velocity during training, whereas the actor selects actions without velocity inputs.

At execution time, each robot estimates the target object and goal landmark positions from onboard RGB images and text inputs using a VLM. We compute text--image similarity over image regions with CLIPSeg~\cite{CLIPSeg} and obtain 2D locations $\hat{x}_t^{\mathrm{obj}}\in\mathbb{R}^2$ by averaging pixel coordinates within high-similarity regions. These pixels are back-projected to 3D world coordinates $\hat{p}_t^{\mathrm{obj}}\in\mathbb{R}^3$ using RGB-D depth and a pinhole camera model with known intrinsics and extrinsics. The same procedure is applied to the goal landmark, and the resulting 3D estimates are used as inputs to the leader--follower policies.


\section{Experiment}
In this section, we evaluate the effectiveness of the proposed method through two simulations and real-robot experiments on cooperative transport with two quadruped robots. The key research questions are as follows:
\begin{itemize}
  \item \textbf{RQ1:} Does CoLF improve training performance? (\ref{subsection:RQ1})
  \item \textbf{RQ2:} Can CoLF maintain consistent cooperative behaviors without goal information for the follower? (\ref{subsection:RQ2})
  \item \textbf{RQ3:} Can CoLF perform effectively with ambiguous language instructions for multiple goal landmarks? (\ref{subsection:RQ3})
  \item \textbf{RQ4:} Can CoLF be effective on real robots and generalize to various goal landmarks? (\ref{subsection:RQ4})
\end{itemize}

\subsection{Experimental Setup}
\label{sim-setup}
\subsubsection{\bf{Simulation Setup}}
All simulation experiments were conducted in a two-robot cooperative transport environment built in NVIDIA Isaac Sim. To accelerate training, we employed 4,096 parallel environments in Isaac Lab.

In this evaluation, we use a single fixed target object, while varying the goal landmark across different scenarios.
The target object is a box with size 0.59~m $\times$ 0.59~m $\times$ 0.35~m and mass 11~kg. The goal landmark is a cylinder with radius 0.3~m and height 0.8~m for evaluation, whereas a point goal is used for training. The target object observation $o_t^{\mathrm{obj}} \in \mathbb{R}^2$ represents the position of the target object relative to the robot, and the goal-landmark observation $o_t^{\mathrm{goal}} \in \mathbb{R}^2$ represents the relative position of the goal landmark. The remaining observation $\tilde{o}_t^{\mathrm{L}} \in \mathbb{R}^9$ consists of the robot’s linear velocity expressed in the robot base frame, the gravity vector expressed in the robot frame, and the relative planar position and yaw angle of the other robot. The action consists of planar linear velocity commands in the robot base frame and an angular velocity command around the vertical axis.

The reward is designed to promote efficient cooperative transport. At each time step $t$, the reward for robot $i\in{\mathrm{L},\mathrm{F}}$ is defined as $r_t^{i}=r_t^{i,\mathrm{rob}}+r_t^{\mathrm{obj}}+r_t^{\mathrm{term}}$.
To maintain proximity to the target object and align the robot heading toward it, we introduce the robot--object proximity reward $r_t^{i,\mathrm{rob}} = w^{i}\exp\bigl(-\lVert p_t^{i}-p_t^{\mathrm{obj}}\rVert\bigr)\bigl(\cos(\theta_t^{i})+0.2\bigr)$, where $p_t^{i}$ and $p_t^{\mathrm{obj}}$ denote the positions of robot $i$ and the target object, and $\theta_t^{i}$ is the angle between the robot yaw direction and the displacement vector $p_t^{\mathrm{obj}}-p_t^{i}$. We set $w^{\mathrm{L}}=2.5$ and $w^{\mathrm{F}}=3.0$ through empirical tuning.
To transport the object toward the goal, we use the object--goal proximity reward $r_t^{\mathrm{obj}} = 6.0 \exp\bigl(-\lVert p_t^{\mathrm{obj}} - p^{\mathrm{goal}}\rVert\bigr)$.
A termination penalty $r_t^{\mathrm{term}}=-2.0$ is applied when a robot falls due to unstable locomotion or collision with the other robot.

\subsubsection{\bf Baselines}
To evaluate the effectiveness of the proposed method ({\bf CoLF}), comparisons were made with the following baseline approaches.

\begin{itemize}
    \item {\bf MAPPO}: A MAPPO baseline where both robots condition their policies on the target object and goal landmark.

    \item {\bf MAPPO (AAC)}: This MAPPO baseline adopts the asymmetric actor--critic (AAC), where the critic is additionally provided with the target object velocity during training.

    \item {\bf CoLF w/o AAC}: To assess the effect of AAC in {\bf CoLF}, we evaluate an ablation where the critic is not provided with the target object velocity during training.

    \item {\bf CoLF w/o CE}: To assess the effect of the CE loss, we evaluate an ablation of {\bf CoLF} without $\mathcal{L}_{\mathrm{CE}}$ in Eq.~(\ref{eq:CEloss}).
\end{itemize}

Note that the asymmetric policy design in {\bf CoLF} refers to role differentiation between the leader and follower policies, which is distinct from the AAC design that introduces information asymmetry between the actor and critic networks. 

All methods are trained with vectorized object and goal positions and executed using VLM-estimated positions. We set $\lambda=0.03$ in all experiments, determined by empirical tuning based on the object--goal proximity reward.

We use a common network architecture across methods: MLP-based actor and critic networks with fully connected layers of sizes 256, 256, and 128, ReLU activations, and linear output layers; they differ only in input dimensionality due to the leader--follower structure.

\subsubsection{\bf Evaluation Metrics}
We evaluate the performance of each method using the following metrics:
\begin{itemize}
    \item {\bf Success Rate (SR):} The percentage of trials in which the final object--goal distance is below a given threshold $\delta$.
    \item {\bf Object--Goal Distance (OGD):} The Euclidean distance between the target object and the goal landmark at the end of each trial.
\end{itemize}

For each method, we evaluate three independently trained policies with different seeds by running 100 trials per seed.

\subsubsection{\bf{VLM Setup}}


At execution time, each robot performs VLM inference conditioned on the text inputs using onboard RGB images.
The onboard camera captures RGB--D images at $1280\times720$ with horizontal and vertical fields of view of $1.5$~rad and $1.0$~rad, respectively, and is mounted $0.40$~m above the robot base frame with a fixed forward-facing orientation.

We compute a text--image similarity map using CLIPSeg with RGB images resized to $224\times224$. The similarity map is thresholded at $0.5$, and the centroid of the high-similarity region is computed as a 2D image location. This location is back-projected to 3D using depth measurements, a pinhole camera model, and known camera intrinsics and extrinsics. Depth values are clipped to $[0.1,10.0]$~m to reflect the range limits of onboard cameras used in real robot experiments.

\subsection{RQ1: Does CoLF Improve Training Performance?}
\label{subsection:RQ1}
\subsubsection{\bf Setup}
We conducted three training runs per method with different random seeds, using the same seeds across methods. The training setup and simulation conditions, including the initial positions of the robots, target object, and goal landmark, are described in Sec.~\ref{sim-setup}.

\subsubsection{\bf Result}

\begin{figure}[t]
  \centering
  \includegraphics[width=0.9\linewidth]{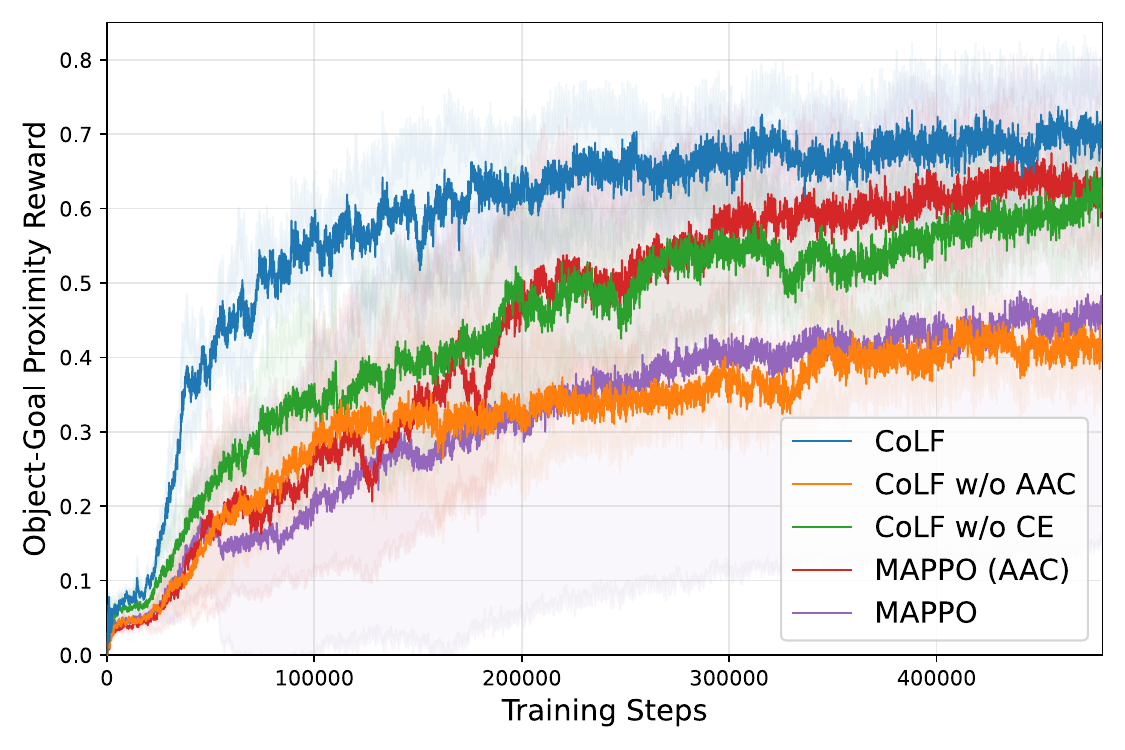}
  \caption{Comparison of the object--goal proximity reward}
  \label{fig:lcurve}
\end{figure}

Figure~\ref{fig:lcurve} shows the learning curves in terms of the object--goal proximity reward $r_t^{\mathrm{obj}}$. Among the MAPPO baselines, \textbf{MAPPO (AAC)} consistently converges to higher reward values than \textbf{MAPPO}, indicating that conditioning the critic on the target-object velocity improves training performance. Notably, \textbf{CoLF} converges to higher reward values than \textbf{MAPPO (AAC)} even though \textbf{MAPPO (AAC)} is trained with the same goal for both robots. This suggests that the CE loss, which encourages the follower to predict the leader’s actions, contributes to improved training performance.

We further evaluate these components in the proposed method. \textbf{CoLF} outperforms \textbf{CoLF w/o AAC}, confirming the contribution of AAC. In addition, compared to \textbf{CoLF w/o CE}, \textbf{CoLF} converges faster and achieves higher reward values, indicating that the CE loss improves training performance.

In summary, these results demonstrate that CoLF is effective during training.

\subsection{RQ2: Can CoLF Maintain Consistent Cooperative Behaviors without Goal Information for the Follower?}
\label{subsection:RQ2}
\subsubsection{\bf Setup}
We assess the effectiveness of CoLF through one-goal-landmark cooperative transport, focusing on maintaining consistent cooperative behaviors without goal information for the follower. The leader and the follower are initialized with $p_0^{\mathrm{L}}\in(-3.0,-2.0)\times(-2.5,-1.0)$ and yaw $-1.5$~rad, and $p_0^{\mathrm{F}}\in(-3.0,-2.0)\times(1.0,2.5)$ and yaw $1.5$~rad, respectively.
The goal landmark position is fixed at $p^{\mathrm{goal}}=(0.5,0.0)$.
We provide the text inputs \emph{``Red box''} for the target object and \emph{``Green cylinder''} for the goal landmark.

\subsubsection{\bf Results}
Table~\ref{tab:result:RQ2} shows quantitative results in one-goal-landmark cooperative transport simulation.
\textbf{CoLF} achieves the highest SR across all methods and attains the smallest OGD despite the absence of goal information for the follower, whereas \textbf{MAPPO} and \textbf{MAPPO (AAC)} provide both robots with goal information.

\begin{figure*}[t]
    \centering
    \includegraphics[width=0.5\textwidth]{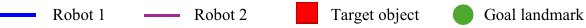}

    \vspace{0.2em}

    \begin{subfigure}{\textwidth}
        \centering
        \begin{subfigure}{0.21\linewidth}
            \centering
            \includegraphics[width=\linewidth]{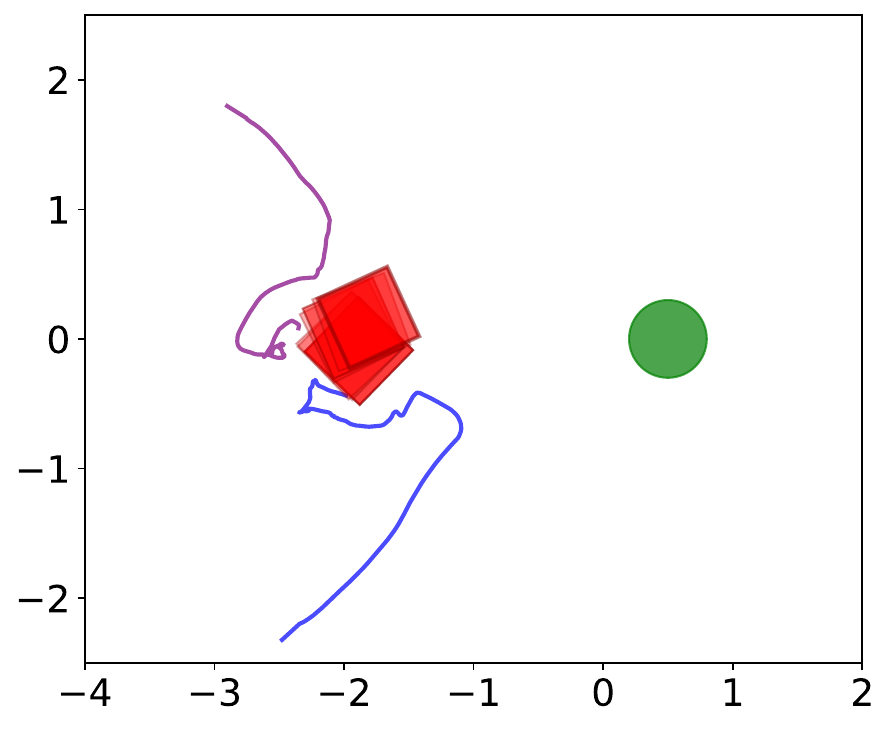}
            \caption*{\bf MAPPO}
        \end{subfigure}\hfill
        \begin{subfigure}{0.21\linewidth}
            \centering
            \includegraphics[width=\linewidth]{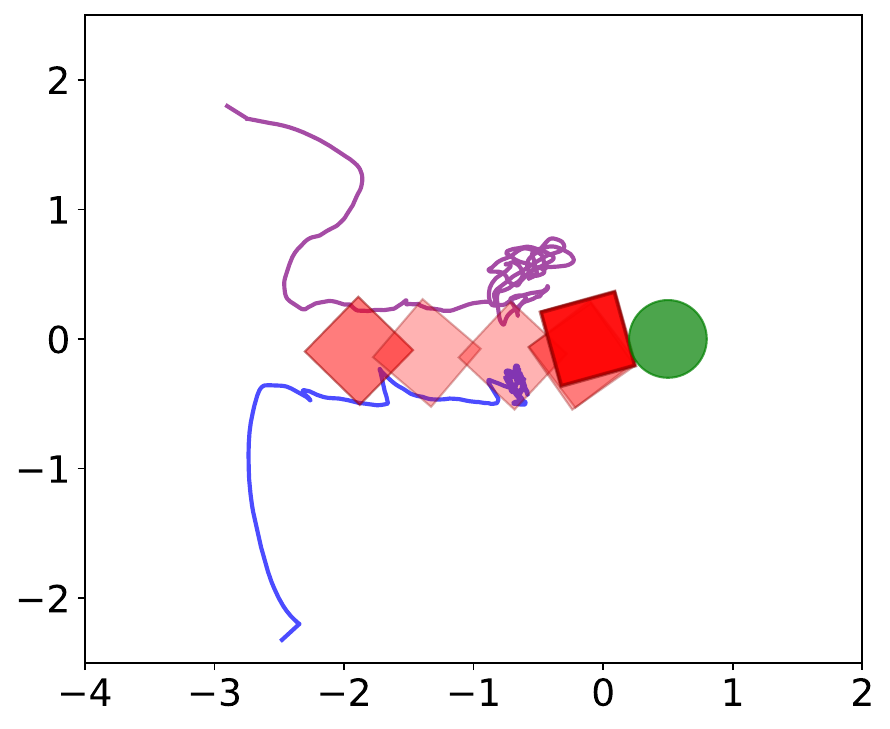}
            \caption*{\bf MAPPO (AAC)}
        \end{subfigure}\hfill
        \begin{subfigure}{0.21\linewidth}
            \centering
            \includegraphics[width=\linewidth]{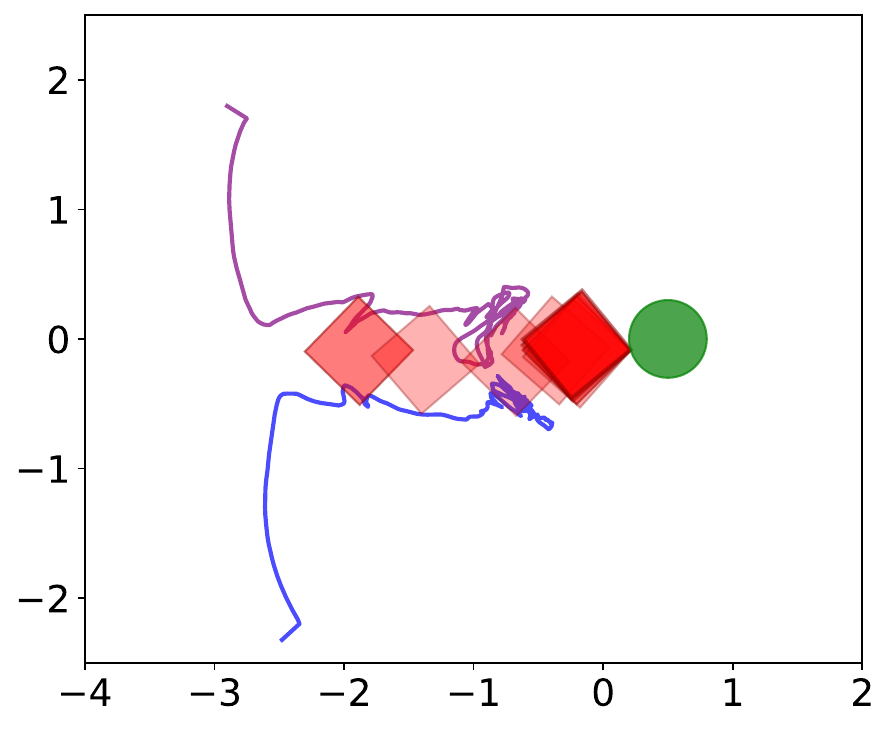}
            \caption*{\bf CoLF w/o CE}
        \end{subfigure}\hfill
        \begin{subfigure}{0.21\linewidth}
            \centering
            \includegraphics[width=\linewidth]{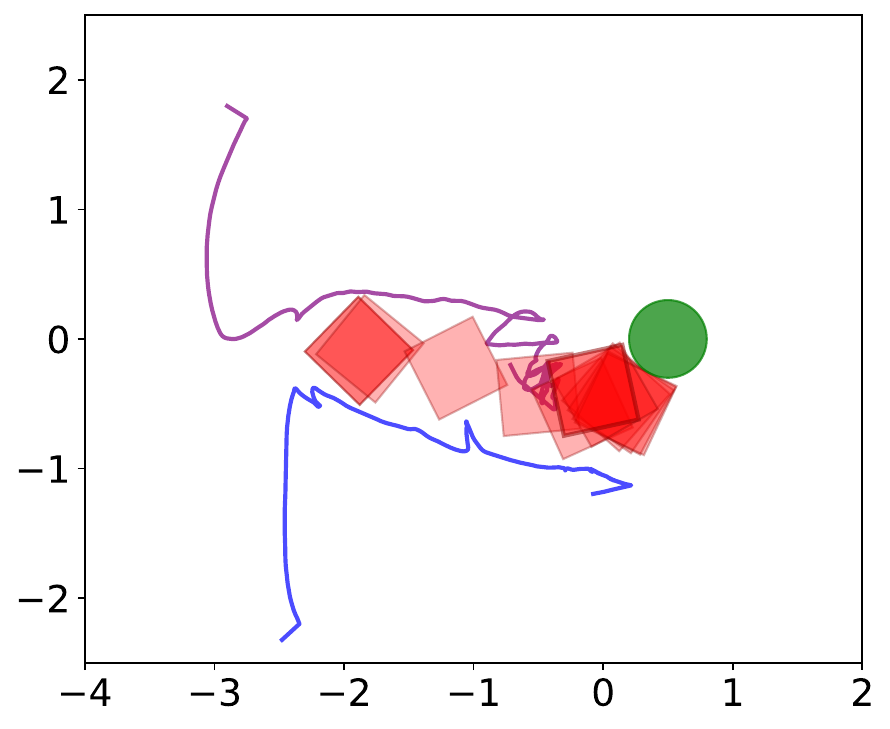}
            \caption*{\bf CoLF}
        \end{subfigure}
        \caption{One-goal-landmark cooperative transport}
        \label{fig:sim1-traj}
    \end{subfigure}

    \vspace{0.5em}

    \begin{subfigure}{\textwidth}
        \centering
        \begin{subfigure}{0.21\linewidth}
            \centering
            \includegraphics[width=\linewidth]{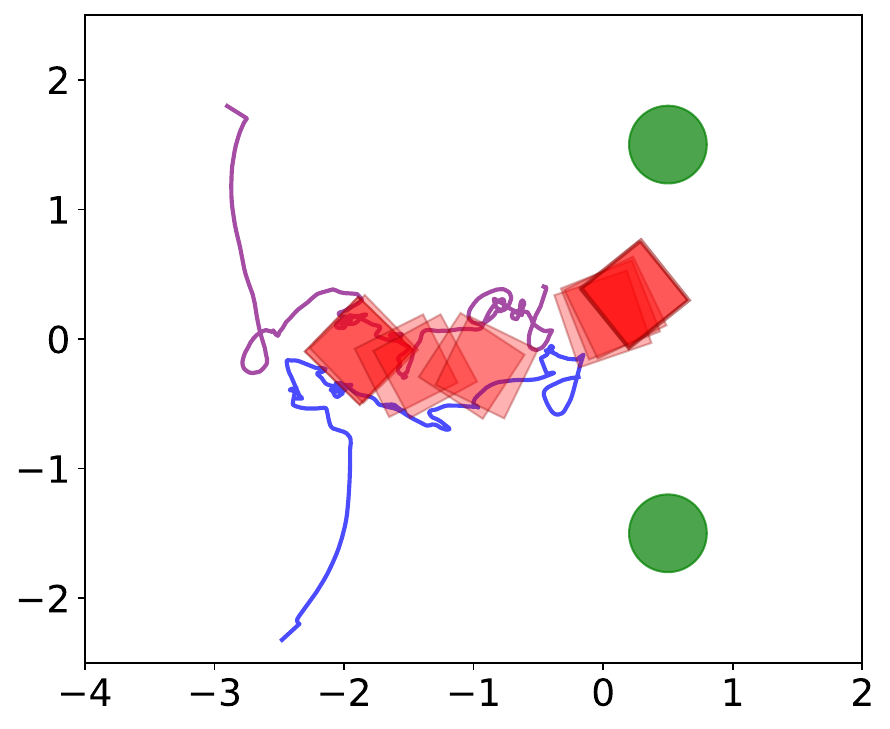}
            \caption*{\bf MAPPO}
        \end{subfigure}\hfill
        \begin{subfigure}{0.21\linewidth}
            \centering
            \includegraphics[width=\linewidth]{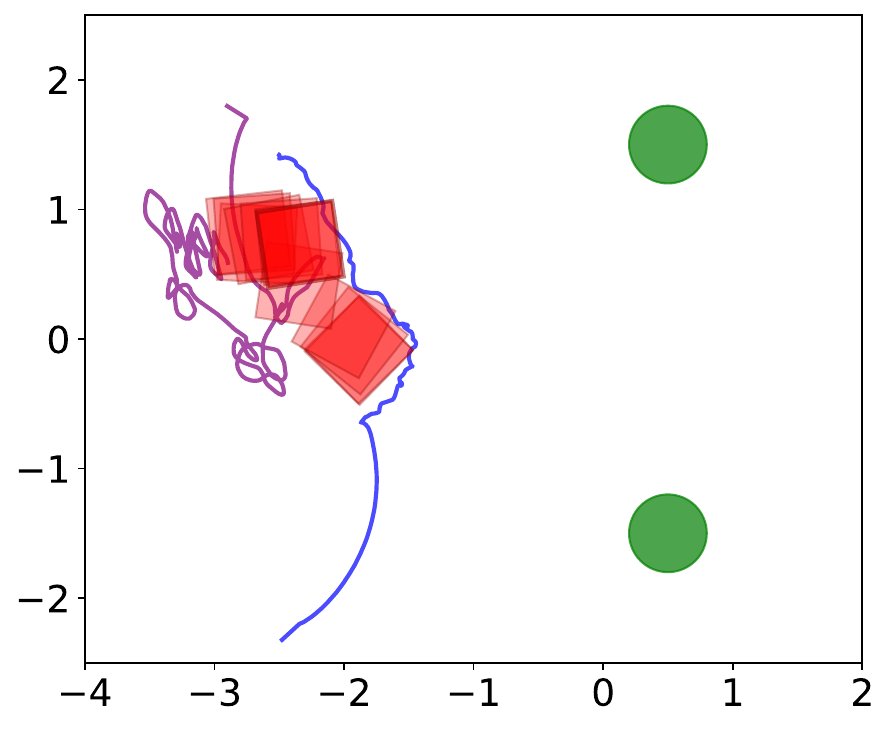}
            \caption*{\bf MAPPO (AAC)}
        \end{subfigure}\hfill
        \begin{subfigure}{0.21\linewidth}
            \centering
            \includegraphics[width=\linewidth]{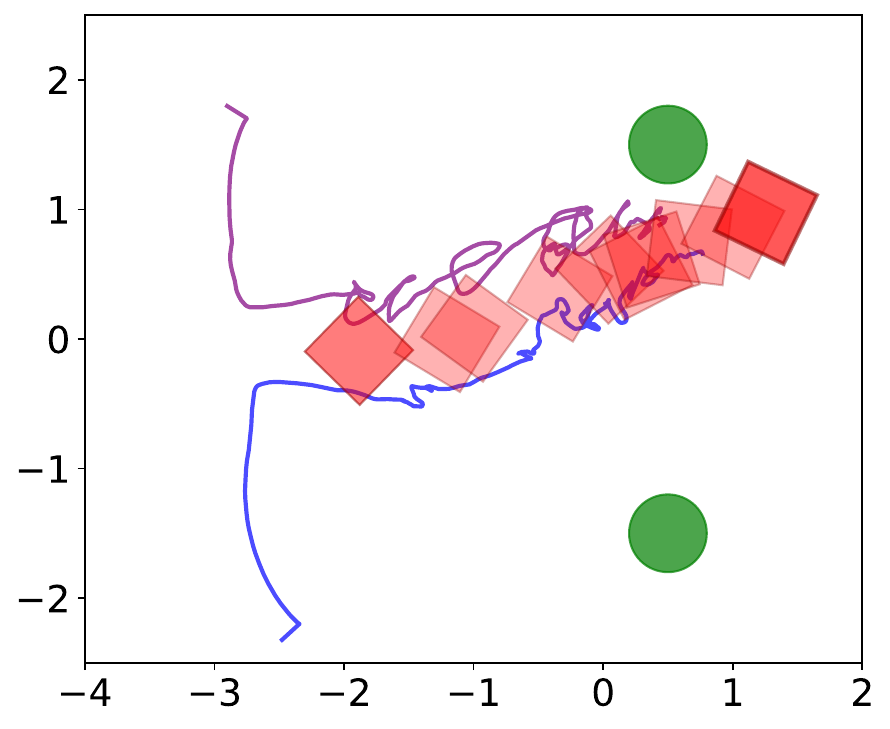}
            \caption*{\bf CoLF w/o CE}
        \end{subfigure}\hfill
        \begin{subfigure}{0.21\linewidth}
            \centering
            \includegraphics[width=\linewidth]{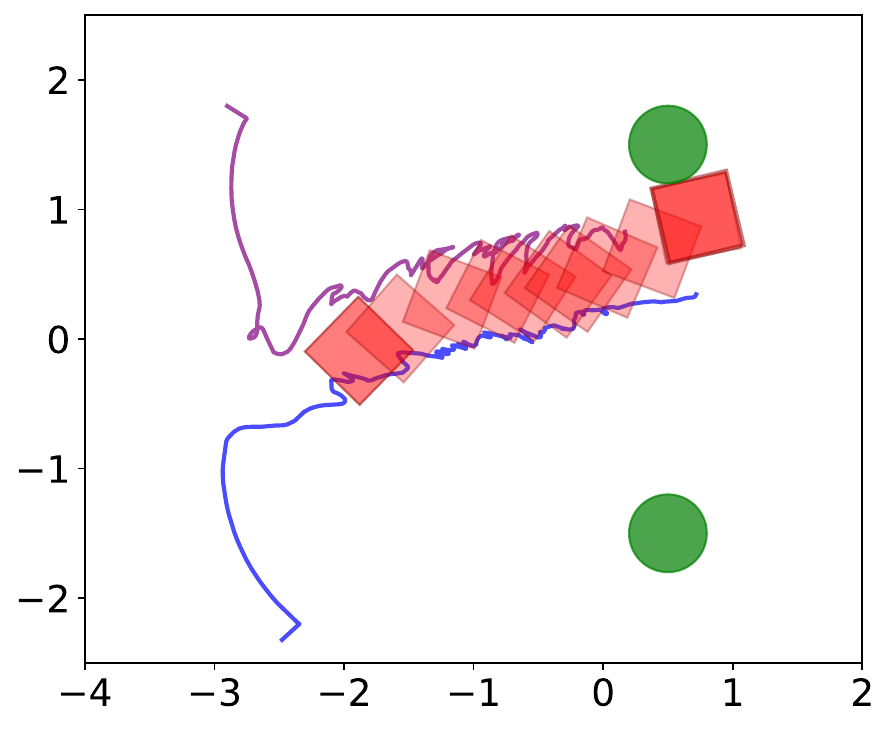}
            \caption*{\bf CoLF}
        \end{subfigure}
        \caption{Two-goal-landmark cooperative transport}
        \label{fig:sim2-traj}
    \end{subfigure}

    \caption{Comparisons of trajectories in two multi-robot cooperative transport scenarios. For \textbf{CoLF w/o CE} and \textbf{CoLF}, Robot~1 and Robot~2 correspond to the Leader and the Follower agents, respectively.}
    \label{fig:traj}
\end{figure*}

Fig.~\ref{fig:sim1-traj} compares trajectories in one-goal-landmark scenarios with identical robots and object initializations. Due to space limitations, we omit \textbf{CoLF w/o AAC}, which showed the lowest performance among the ablated variants of \textbf{CoLF}. We observe that \textbf{MAPPO (AAC)}, \textbf{CoLF w/o CE}, and \textbf{CoLF} successfully transport the target object to the goal landmark, whereas \textbf{MAPPO} fails to reach the goal landmark.

Overall, the proposed method can maintain consistent cooperative behaviors without goal information for the follower.

\begin{table}[t]
\centering
\caption{Comparisons in one-goal-landmark cooperative transport simulation. The minimum achievable $\delta$ and OGD are $0.59$~m, attained when the target object is tangent to the cylindrical goal landmark.}
\label{tab:result:RQ2}
\begin{tabular}{lccc}
\toprule
Method & SR (\(\delta<0.65\) m) & SR (\(\delta<0.75\) m) & OGD [m] \\
\midrule
\textbf{MAPPO}         & 0.18 ± 0.12  & 0.29 ± 0.19  & 1.39 ± 0.37  \\
\textbf{MAPPO (AAC)}   & 0.35 ± 0.06  & 0.71 ± 0.21  & 1.02 ± 0.26  \\
\textbf{CoLF w/o AAC}  & 0.07 ± 0.07   & 0.16 ± 0.16  & 1.49 ± 0.43  \\
\textbf{CoLF w/o CE}   & 0.33 ± 0.20  & 0.55 ± 0.31   & 0.99 ± 0.28  \\
\textbf{CoLF}          & \bf{0.44 ± 0.09}   & \bf{0.76 ± 0.17}   & \bf{0.80 ± 0.17}   \\
\bottomrule
\end{tabular}
\end{table}

\subsection{RQ3: Can CoLF Perform Effectively with Ambiguous Language Instructions for Multiple Goal Landmarks?}
\label{subsection:RQ3}

\subsubsection{\bf Setup}
We further assess the effectiveness of CoLF through two-goal-landmark cooperative transport, focusing on maintaining consistent cooperative behaviors under ambiguous language instructions in the presence of multiple similar goal landmarks.
Two goal landmarks are placed at fixed positions,
$p^{\mathrm{goal}}_1=(0.5,-1.5)$ and $p^{\mathrm{goal}}_2=(0.5,1.5)$.
All other experimental settings, including the text inputs, are identical to those used in the one-goal-landmark cooperative transport. We evaluate each method by deploying a policy trained only in a single-goal setting, without any additional training.

\subsubsection{\bf Results}
Table~\ref{tab:result:RQ3} shows quantitative results in the two-goal-landmark simulation.
\textbf{CoLF} achieves the highest SR and the lowest OGD, while all baselines exhibit lower success rates, indicating that goal misalignment across robots is critical when multiple similar landmarks are present.

Fig.~\ref{fig:sim2-traj} visualizes representative trajectories under identical robot and target object initializations.
The results show that \textbf{CoLF} successfully transports the target object toward the correct goal landmark, whereas the other methods fail to transport the target object to the goal landmark.

Overall, the proposed method maintains consistent cooperative behaviors  under ambiguous language instructions in the presence of multiple goal landmarks.

\begin{table}[t]
\centering
\caption{Comparisons in two-goal-landmark cooperative transport simulation}
\label{tab:result:RQ3}
\begin{tabular}{lccc}
\toprule
Method & SR (\(\delta<0.65\) m) & SR (\(\delta<0.75\) m) & OGD [m] \\
\midrule
\textbf{MAPPO}         & 0.09 ± 0.10   & 0.14 ± 0.12   & 1.58 ± 0.35   \\
\textbf{MAPPO (AAC)}   & 0.17 ± 0.04 & 0.29 ± 0.08 & 1.31 ± 0.12 \\
\textbf{CoLF w/o AAC}  & 0.04 ± 0.02   & 0.09 ± 0.05   & 1.65 ± 0.37   \\
\textbf{CoLF w/o CE}   & 0.22 ± 0.11 & 0.34 ± 0.14 & 1.19 ± 0.32 \\
\textbf{CoLF}          & \bf{0.59 ± 0.17} & \bf{0.79 ± 0.08} & \bf{0.78 ± 0.09} \\
\bottomrule
\end{tabular}
\end{table}

\subsection{RQ4: Can CoLF be Effective on Real Robots and Generalize to Various Goal Landmarks?}
\label{subsection:RQ4}

\subsubsection{\bf Setup}
We evaluate the proposed method in real-world experiments using two Unitree Go1 quadruped robots. Each robot is equipped with an RGB--D camera, Intel RealSense D455, mounted at a height of 0.55~m. A motion-capture system is used to measure the relative pose between the robots. In these experiments, we use a single target object with fixed shape, mass, and friction. 

\begin{figure*}[t]
    \centering
    \begin{subfigure}{0.19\textwidth}
        \centering
        \includegraphics[width=\linewidth]{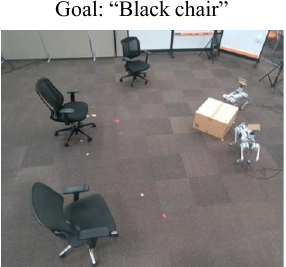}
        \caption{Multi-Identical Chairs}
    \end{subfigure}\hfill
    \begin{subfigure}{0.19\textwidth}
        \centering
        \includegraphics[width=\linewidth]{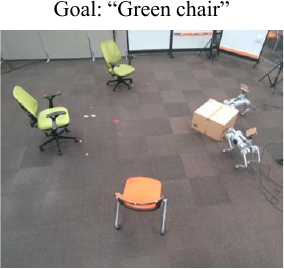}
        \caption{Multi-Colored-Chairs}
    \end{subfigure}\hfill
    \begin{subfigure}{0.19\textwidth}
        \centering
        \includegraphics[width=\linewidth]{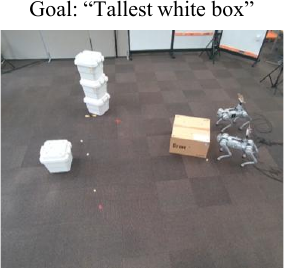}
        \caption{Different-Height-Boxes}
    \end{subfigure}\hfill
    \begin{subfigure}{0.19\textwidth}
        \centering
        \includegraphics[width=\linewidth]{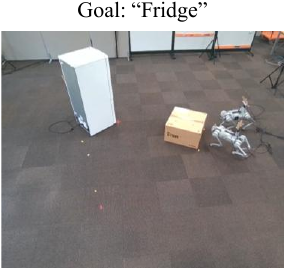}
        \caption{Single-Fridge}
    \end{subfigure}\hfill
    \begin{subfigure}{0.19\textwidth}
        \centering
        \includegraphics[width=\linewidth]{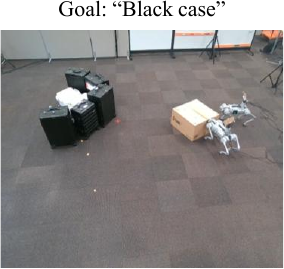}
        \caption{Clustered-Cases}
    \end{subfigure}

    \caption{Overview of the five real-world scenarios.}
    \label{fig:scenario}
\end{figure*}

We compare \textbf{CoLF} with \textbf{MAPPO (AAC)}, the best-performing MAPPO baseline in simulation.
To evaluate real-world generalization, we conduct experiments in five scenarios with different goal landmarks, as shown in Fig.~\ref{fig:scenario}.
Each scenario includes five trials with different initial configurations of the robots, the target object, and the goal landmark, shared across methods.
A trial is counted as successful if the final object--landmark distance is at most 0.20~m larger than the minimum achievable distance.

\subsubsection{\bf Results}
Table~\ref{tab:real_results} shows real-world success rates in five scenarios.
Both \textbf{MAPPO (AAC)} and \textbf{CoLF} achieve high success rates in Different-Height-Boxes and Single-Fridge, where goal misalignment is less likely to arise.
In Multi-Identical-Chairs, Multi-Colored-Chairs, and Clustered-Boxes, where goal misalignment is expected to occur, \textbf{CoLF} consistently achieves higher success rates than \textbf{MAPPO (AAC)}.
See the attached video for all scenarios.

Overall, the proposed method is effective on real robots and generalizes to various goal landmarks.

\begin{table}[t]
\centering
\caption{Real-world success rate over five trials in five scenarios}
\label{tab:real_results}
\begin{tabular}{lcc}
\toprule
Task Scenario & \textbf{MAPPO (AAC)} & \textbf{CoLF} \\
\midrule
Multi-Identical-Chairs    & 2/5 & \bf{5/5} \\
Multi-Colored-Chairs      & 1/5 & \bf{5/5} \\
Different-Height-Boxes    & 3/5 & \bf{5/5} \\
Single-Fridge       & 3/5 & \bf{4/5} \\
Clustered-Cases           & 2/5 & \bf{5/5} \\
\bottomrule
\end{tabular}
\end{table}

\section{Discussion}We demonstrated the effectiveness of the proposed method in real-robot experiments using a single object with fixed shape, mass, and friction.
However, the control performance can degrade under variations in object properties.
To address this issue, we will incorporate domain randomization during training~\cite{DomainRandom} and quantitatively evaluate the policies in real-robot experiments under diverse object conditions.

Our current study assumes that each robot has access to the other robot’s relative position and yaw angle to avoid inter-robot collisions. While this information could, in principle, be inferred from onboard visual observations, reliable pose estimation during object transport requires active viewpoint control to maintain visibility of the other robot. To address this issue, future work will integrate the proposed method with collision-aware active viewpoint control and evaluate its effectiveness in a fully decentralized setting.

Scalability to team-size changes is an important direction for future work. In this study, we focus on the effects of viewpoint discrepancies and language ambiguity on cooperative behavior under individual VLM inference, using a minimal two-robot setup. However, since we adopt MAPPO, the policy input dimensions depend on the number of robots, which hinders deployment to teams of different sizes. To address this limitation, future work should combine the proposed approach with decentralized MARL methods that are invariant to the number of robots~\cite{Shibata2023RAS,Chen2025} and evaluate scalability on real robots.

In this study, we assume that the leader can correctly recognize the target object and the goal landmark during execution. However, the inferred target or goal may differ from the user’s intent under ambiguous language instructions. To address this issue, we plan to combine language instructions with other instruction modalities, such as human pointing~\cite{Deguchi2023}, to better reflect the user’s intent during execution.

\section{Conclusion}
This paper presented CoLF, a MARL framework for vision--language-guided multi-robot cooperative transport under decentralized perception. CoLF promotes stable leader--follower role differentiation through two key components: (1) an asymmetric policy design and (2) a mutual-information-based training objective that maximizes a variational lower bound between the leader’s action and the follower’s predicted leader action. Simulation results showed that CoLF achieves higher success rates than baseline methods under perceptual misalignment conditions. Real-robot experiments with two quadruped robots across five scenarios further demonstrated successful sim-to-real transfer and generalization to various goal landmarks. As future work, we will evaluate scalability across varying team sizes and extend the framework to more challenging tasks, including 3D cooperative manipulation.

\section*{Acknowledgments}
This work was supported by the Tateishi Science and Technology Foundation through Research Grant (A).






\bibliographystyle{IEEEtran}
\bibliography{IEEEexample}             

@ARTICLE{Koung2021,
  author={Koung, Daravuth and Kermorgant, Olivier and Fantoni, Isabelle and Belouaer, Lamia},
  journal={IEEE Robotics and Automation Letters}, 
  title={Cooperative Multi-Robot Object Transportation System Based on Hierarchical Quadratic Programming}, 
  year={2021},
  volume={6},
  number={4},
  pages={6466-6472}
}

@ARTICLE{Hu2022,
  author={Hu, Jiawei and Liu, Wenhang and Zhang, Heng and Yi, Jingang and Xiong, Zhenhua},
  journal={IEEE Robotics and Automation Letters}, 
  title={Multi-Robot Object Transport Motion Planning With a Deformable Sheet}, 
  year={2022},
  volume={7},
  number={4},
  pages={9350-9357}
}

@ARTICLE{Kennel2024,
  author={Kennel-Maushart, Florian and Coros, Stelian},
  journal={IEEE Robotics and Automation Letters}, 
  title={Payload-Aware Trajectory Optimisation for Non-Holonomic Mobile Multi-Robot Manipulation With Tip-Over Avoidance}, 
  year={2024},
  volume={9},
  number={9},
  pages={7669-7676}
}

@INPROCEEDINGS{Li2023,
  author={Li, Guanrui and Loianno, Giuseppe},
  booktitle={2023 IEEE/RSJ International Conference on Intelligent Robots and Systems (IROS)}, 
  title={Nonlinear Model Predictive Control for Cooperative Transportation and Manipulation of Cable Suspended Payloads with Multiple Quadrotors}, 
  year={2023},
  pages={5034-5041}
}

@article{Sihao2025,
  author = {Sihao Sun  and Xuerui Wang  and Dario Sanalitro  and Antonio Franchi  and Marco Tognon  and Javier Alonso-Mora },
  title = {Agile and cooperative aerial manipulation of a cable-suspended load},
  journal = {Science Robotics},
  volume = {10},
  number = {107},
  pages = {eadu8015},
  year = {2025}
}

@INPROCEEDINGS{Fawcett2023,
  author={Fawcett, Randall T. and Amanzadeh, Leila and Kim, Jeeseop and Ames, Aaron D. and Hamed, Kaveh Akbari},
  booktitle={2023 IEEE International Conference on Robotics and Automation (ICRA)}, 
  title={Distributed Data-Driven Predictive Control for Multi-Agent Collaborative Legged Locomotion}, 
  year={2023},
  pages={9924-9930}
}

@ARTICLE{Carli2025,
  author={De Carli, Nicola and Belletti, Riccardo and Buzzurro, Emanuele and Testa, Andrea and Notarstefano, Giuseppe and Tognon, Marco},
  journal={IEEE Robotics and Automation Letters}, 
  title={Distributed NMPC for Cooperative Aerial Manipulation of Cable-Suspended Loads}, 
  year={2025},
  volume={10},
  number={10},
  pages={10546-10553}
}

@ARTICLE{Culbertson2021,
  author={Culbertson, Preston and Slotine, Jean-Jacques and Schwager, Mac},
  journal={IEEE Transactions on Robotics}, 
  title={Decentralized Adaptive Control for Collaborative Manipulation of Rigid Bodies}, 
  year={2021},
  volume={37},
  number={6},
  pages={1906-1920}
}

@ARTICLE{Yan2021,
  author={Yan, Lei and Stouraitis, Theodoros and Vijayakumar, Sethu},
  journal={IEEE Robotics and Automation Letters}, 
  title={Decentralized Ability-Aware Adaptive Control for Multi-Robot Collaborative Manipulation}, 
  year={2021},
  volume={6},
  number={2},
  pages={2311-2318}
}

@inproceedings{Wang2016,
  title={Kinematic multi-robot manipulation with no communication using force feedback},
  author={Wang, Zijian and Schwager, Mac},
  booktitle={2016 IEEE international conference on robotics and automation (ICRA)},
  pages={427--432},
  year={2016},
  organization={IEEE}
}

@INPROCEEDINGS{Mehta2025,
  author={Mehta, Dhruv K and Joglekar, Ajinkya and Krovi, Venkat},
  booktitle={2025 IEEE International Conference on Robotics and Automation (ICRA)}, 
  title={Deep Reinforcement Learning for Coordinated Payload Transport in Biped-Wheeled Robots}, 
  year={2025},
  pages={14992-14998}
}

@INPROCEEDINGS{Feng2025,
  author={Feng, Yuming and Hong, Chuye and Niu, Yaru and Liu, Shiqi and Yang, Yuxiang and Zhao, Ding},
  booktitle={2025 IEEE International Conference on Robotics and Automation (ICRA)}, 
  title={Learning Multi-Agent Loco-Manipulation for Long-Horizon Quadrupedal Pushing}, 
  year={2025},
  pages={14441-14448},
}

@InProceedings{Pandit2024,
  title = 	 {Learning Decentralized Multi-Biped Control for Payload Transport},
  author =       {Pandit, Bikram and Gupta, Ashutosh and Gadde, Mohitvishnu S. and Johnson, Addison and Shrestha, Aayam Kumar and Duan, Helei and Dao, Jeremy and Fern, Alan},
  booktitle = 	 {Proceedings of The 8th Conference on Robot Learning},
  pages = 	 {1021--1034},
  year = 	 {2025},
  volume = 	 {270},
  publisher =    {PMLR}
}

@InProceedings{Zeng2025,
  title = 	 {Decentralized Aerial Manipulation of a Cable-Suspended Load Using Multi-Agent Reinforcement Learning},
  author =       {Zeng, Jack and Gimenez, Andreu Matoses and Vinitsky, Eugene and Alonso-Mora, Javier and Sun, Sihao},
  booktitle = 	 {Proceedings of The 9th Conference on Robot Learning},
  pages = 	 {3850--3868},
  year = 	 {2025},
  volume = 	 {305},
  publisher =    {PMLR}
}

@article{Shibata2023RAS,
title = {Deep reinforcement learning of event-triggered communication and consensus-based control for distributed cooperative transport},
journal = {Robotics and Autonomous Systems},
volume = {159},
pages = {104307},
year = {2023},
author = {Kazuki Shibata and Tomohiko Jimbo and Takamitsu Matsubara}
}

@article{Chen2025,
  title={Decentralized Navigation of a Cable-Towed Load using Quadrupedal Robot Team via MARL},
  author={Chen, Wen-Tse and Nguyen, Minh and Li, Zhongyu and Sue, Guo Ning and Sreenath, Koushil},
  journal={arXiv preprint arXiv:2503.18221},
  year={2025}
}

@article{CollaBot,
  title={{CollaBot}: Vision-language guided simultaneous collaborative manipulation},
  author={Song, Kun and Ma, Shentao and Chen, Gaoming and Jin, Ninglong and Zhao, Guangbao and Ding, Mingyu and Xiong, Zhenhua and Pan, Jia},
  journal={arXiv preprint arXiv:2508.03526},
  year={2025}
}

@inproceedings{VIKI-R,
  title={{VIKI-R}: Coordinating Embodied Multi-Agent Cooperation via Reinforcement Learning},
  author={Li Kang and Xiufeng Song and Heng Zhou and Yiran Qin and Jie Yang and Xiaohong Liu and Philip Torr and LEI BAI and Zhenfei Yin},
  booktitle={The Thirty-ninth Annual Conference on Neural Information Processing Systems Datasets and Benchmarks Track},
  year={2025}
}

@inproceedings{COMBO,
  title={{COMBO}: Compositional World Models for Embodied Multi-Agent Cooperation},
  author={Hongxin Zhang and Zeyuan Wang and Qiushi Lyu and Zheyuan Zhang and Sunli Chen and Tianmin Shu and Behzad Dariush and Kwonjoon Lee and Yilun Du and Chuang Gan},
  booktitle={The Thirteenth International Conference on Learning Representations},
  year={2025}
}

@article{HierarchicalLM,
  title={Hierarchical Language Models for Semantic Navigation and Manipulation in an Aerial-Ground Robotic System},
  author={Liu, Haokun and Ma, Zhaoqi and Li, Yunong and Sugihara, Junichiro and Chen, Yicheng and Li, Jinjie and Zhao, Moju},
  journal={Advanced Intelligent Systems},
  pages={e202500640},
  year={2025},
  publisher={Wiley Online Library}
}

@article{TRACE,
  title={{TRACE}: A Self-Improving Framework for Robot Behavior Forecasting with Vision-Language Models},
  author={Puthumanaillam, Gokul and Padrao, Paulo and Fuentes, Jose and Thangeda, Pranay and Schafer, William E and Song, Jae Hyuk and Jagdale, Karan and Bobadilla, Leonardo and Ornik, Melkior},
  journal={arXiv preprint arXiv:2503.00761},
  year={2025}
}

@inproceedings{MCoCoNav,
  title={Enhancing multi-robot semantic navigation through multimodal chain-of-thought score collaboration},
  author={Shen, Zhixuan and Luo, Haonan and Chen, Kexun and Lv, Fengmao and Li, Tianrui},
  booktitle={Proceedings of the AAAI Conference on Artificial Intelligence},
  volume={39},
  number={14},
  pages={14664--14672},
  year={2025}
}

@INPROCEEDINGS{ZeroCAP,
  author={Venkatesh, Vishnunandan L. N. and Min, Byung-Cheol},
  booktitle={2025 IEEE International Conference on Robotics and Automation (ICRA)}, 
  title={{ZeroCAP}: Zero-Shot Multi-Robot Context Aware Pattern Formation via Large Language Models}, 
  year={2025},
  pages={01-07}
}

@INPROCEEDINGS{CLIPSwarm,
  author={Pueyo, Pablo and Montijano, Eduardo and Murillo, Ana C. and Schwager, Mac},
  booktitle={2024 IEEE/RSJ International Conference on Intelligent Robots and Systems (IROS)}, 
  title={{CLIPSwarm}: Generating Drone Shows from Text Prompts with Vision-Language Models}, 
  year={2024},
  pages={11917-11923}
}

@book{Dec-POMDP,
author = {Oliehoek, Frans A. and Amato, Christopher},
title = {A Concise Introduction to Decentralized POMDPs},
year = {2016},
publisher = {Springer Publishing Company, Incorporated},
edition = {1st},
}

@inproceedings{mappo,
  title={The surprising effectiveness of {PPO} in cooperative multi-agent games},
  author={Yu, Chao and Velu, Akash and Vinitsky, Eugene and Gao, Jiaxuan and Wang, Yu and Bayen, Alexandre and Wu, Yi},
  booktitle={Advances in Neural Information Processing Systems},
  volume={35},
  pages={24611--24624},
  year={2022}
}

@article{ppo,
  title={Proximal policy optimization algorithms},
  author={Schulman, John and Wolski, Filip and Dhariwal, Prafulla and Radford, Alec and Klimov, Oleg},
  journal={arXiv preprint arXiv:1707.06347},
  year={2017}
}

@inproceedings{Omron2024,
  title={Language-Guided Pattern Formation for Swarm Robotics with Multi-Agent Reinforcement Learning},
  author={Liu, Hsu-Shen and Kuroki, So and Kozuno, Tadashi and Sun, Wei-Fang and Lee, Chun-Yi},
  booktitle={2024 IEEE/RSJ International Conference on Intelligent Robots and Systems (IROS)},
  pages={8998--9005},
  year={2024},
}

@INPROCEEDINGS{Yano2025,
  author={Yano, Yoshiki and Shibata, Kazuki and Kokshoorn, Maarten and Matsubara, Takamitsu},
  booktitle={2025 IEEE/RSJ International Conference on Intelligent Robots and Systems (IROS)}, 
  title={ICCO: Learning an Instruction-conditioned Coordinator for Language-guided Task-aligned Multi-robot Control}, 
  year={2025},
  volume={},
  number={},
  pages={18700-18707}
}

@inproceedings{asymmetric-actor-critic,
  title={Asymmetric Actor Critic for Image-Based Robot Learning},
  author={Lerrel Pinto and Marcin Andrychowicz and Peter Welinder and Wojciech Zaremba and Pieter Abbeel},
  booktitle={Robotics: Science and Systems (RSS)},
  year={2018}
}

@InProceedings{CLIPSeg,
    author    = {L\"uddecke, Timo and Ecker, Alexander},
    title     = {Image Segmentation Using Text and Image Prompts},
    booktitle = {Proceedings of the IEEE/CVF Conference on Computer Vision and Pattern Recognition (CVPR)},
    month     = {June},
    year      = {2022},
    pages     = {7086-7096}
}

@InProceedings{DomainRandom,
  title = 	 {Active Domain Randomization},
  author =       {Mehta, Bhairav and Diaz, Manfred and Golemo, Florian and Pal, Christopher J. and Paull, Liam},
  booktitle = 	 {Proceedings of the Conference on Robot Learning},
  pages = 	 {1162--1176},
  year = 	 {2020},
  editor = 	 {Kaelbling, Leslie Pack and Kragic, Danica and Sugiura, Komei},
  volume = 	 {100}
}

@INPROCEEDINGS{Deguchi2023,
  author={Deguchi, Hideki and Taguchi, Shun and Shibata, Kazuki and Koide, Satoshi},
  booktitle={2023 IEEE/RSJ International Conference on Intelligent Robots and Systems (IROS)}, 
  title={Enhanced Robot Navigation with Human Geometric Instruction}, 
  year={2023},
  pages={9071-9078},
}

\end{document}